\documentclass[10pt,twocolumn,letterpaper]{article}

\usepackage{iccv}
\usepackage{times}
\usepackage{epsfig}
\usepackage{graphicx}
\usepackage{amsmath}
\usepackage{amssymb}
\usepackage{subcaption}
\usepackage{comment}
\usepackage{enumitem}
\usepackage{booktabs}
\usepackage{colortbl}
\usepackage{tabularx}
\usepackage{multirow}
\usepackage{balance}
\usepackage{caption}
\usepackage[pagebackref=true,breaklinks=true,colorlinks,bookmarks=false]{hyperref}

\iccvfinalcopy

\begin{document}

\title{ID-Reveal: Identity-aware DeepFake Video Detection}

\author{
	Davide Cozzolino\textsuperscript{1} \ \,\,\,
	Andreas R\"ossler\textsuperscript{2} \ \ \,\,\,
	Justus Thies\textsuperscript{2,3} \ \ \,\,\,
	Matthias Nie{\ss}ner\textsuperscript{2} \ \ \,\,\,
	Luisa Verdoliva\textsuperscript{1} \\ \\
	\textsuperscript{1}University Federico II of Naples \ \ \ \ \ \textsuperscript{2}Technical University of Munich \ \ \ \ \ \\ \textsuperscript{3}Max Planck Institute for Intelligent Systems, Tübingen \\ \\
}

\maketitle

\newcommand{\GenNet}{3DMM Generative Network}
\newcommand{\TemporalIDNetwork}{Temporal ID Network}
\newcommand{\ru}{\rule{0mm}{3mm}}
\newcommand{\rota}[1]{\rotatebox[origin=c]{90}{#1}}

\captionsetup{font=small}

\begin{abstract}

A major challenge in DeepFake forgery detection is that state-of-the-art algorithms are mostly trained to detect a specific fake method.
As a result, these approaches show poor generalization across different types of facial manipulations, e.g., from face swapping to facial reenactment.
To this end, we introduce ID-Reveal, a new approach that learns temporal facial features, specific of how a person moves while talking, by means of metric learning coupled with an adversarial training strategy.
The advantage is that we do not need any training data of fakes, but only train on real videos. 
Moreover, we utilize high-level semantic features, which enables robustness to widespread and disruptive forms of post-processing.
We perform a thorough experimental analysis on several publicly available benchmarks. 
Compared to state of the art, our method improves generalization and is more robust to low-quality videos, that are usually spread over social networks. 
In particular, we obtain an average improvement of more than 15\% in terms of accuracy for facial reenactment on high compressed videos.

\end{abstract}

\section{Introduction}
Recent advancements in synthetic media generation allow us to automatically manipulate images and videos with a high level of realism.
To counteract the misuse of these image synthesis and manipulation methods, the digital media forensics field got a lot of attention \cite{Verdoliva2020media, Tolosana2020DeepFakes}.
For instance, during the past two years, there has been intense research on DeepFake detection, that has been strongly stimulated by the introduction of large datasets of videos with manipulated faces \cite{Roessler2019ff++, Dufour2019, Dolhansky2020dfdc, Li2020celeb, Jiang2020DeeperForensics, Li2020face, Fox2020VideoForensicsHQ}.

However, despite excellent detection performance, the major challenge is how to generalize to previously unseen methods.
For instance, a detector trained on face swapping will drastically drop in performance when tested on a facial reenactment method. 
This unfortunately limits practicality as we see new types of forgeries appear almost on a daily basis.
As a result, supervised detection, which requires extensive training data of a specific forgery method, cannot immediately detect a newly-seen forgery type.
\begin{figure}[t!]
    \centering
    \includegraphics[width=\linewidth]{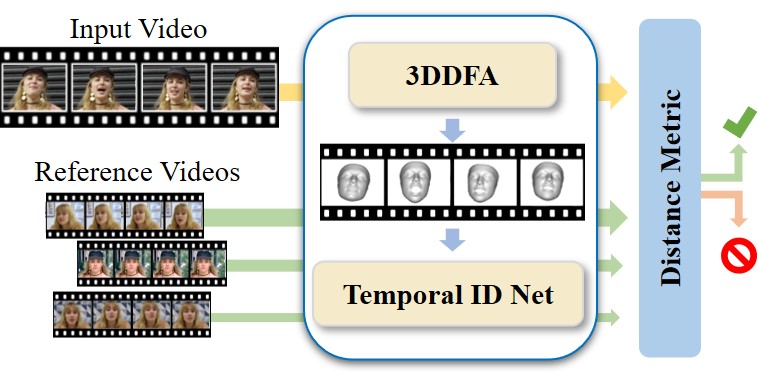}
    \caption{ID-Reveal is an identity-aware DeepFake video detection. Based on reference videos of a person, we estimate a temporal embedding which is used as a distance metric to detect fake videos.}
    \label{fig:proposal_scheme}
\end{figure}

This mismatch and generalization issue has been addressed in the literature using different strategies, ranging from applying domain adaptation \cite{Cozzolino2018, Aneja2020}  or active learning \cite{Du2020towards} to strongly increasing augmentation during training \cite{Wang2020CNN, Dolhansky2020dfdc} or by means of ensemble procedures \cite{Dolhansky2020dfdc, Bonettini2020video}.
A different line of research is relying only on pristine videos at training time and detecting possible anomalies with respect to forged ones \cite{Huh2018, Cozzolino2019extracting, Cozzolino2020}.
This can help to increase the generalization ability with respect to new unknown manipulations but does not solve the problem of videos characterized by a different digital history.
This is quite common whenever a video is spread over social networks and posted multiple times by different users.
In fact, most of the platforms often reduce the quality and/or the video resolution.

\begin{figure}[!t]
    \centering
    \includegraphics[width=0.78\linewidth]{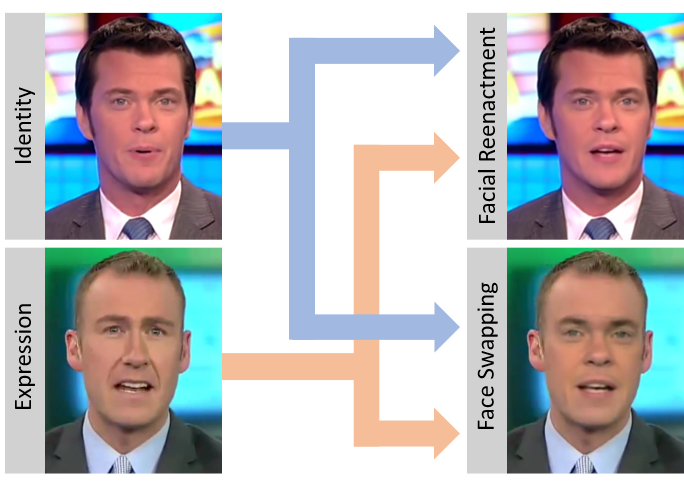}
    \caption{Automatic face manipulations can be split in two main categories: facial reenactment and face-swapping. The first one alters the facial expression preserving the identity. The second one modifies the identity of a person preserving the facial expression.
    }
    \label{fig:intro}
\end{figure}

Note also that current literature has mostly focused on face-swapping, a manipulation that replaces the facial identity of a subject with another one,
however, a very effective modification is facial reenactment~\cite{thies2016face}, where only the expression or the lips movements of a person are modified (Fig.~\ref{fig:intro}).
Recently, the MIT Center for Advanced Virtuality created a DeepFake video of president Richard Nixon \footnote{\url{https://moondisaster.org}}.
The synthetic video shows Nixon giving a speech he never intended to deliver, by modifying only the lips movement and the speech of the old pristine video.
The final result is impressive and shows the importance to develop forgery detection approaches that can generalize on different types of facial manipulations.

\begin{figure}[t!]
    \centering
    \includegraphics[width=0.96\linewidth,trim=3 10 0 0, clip]{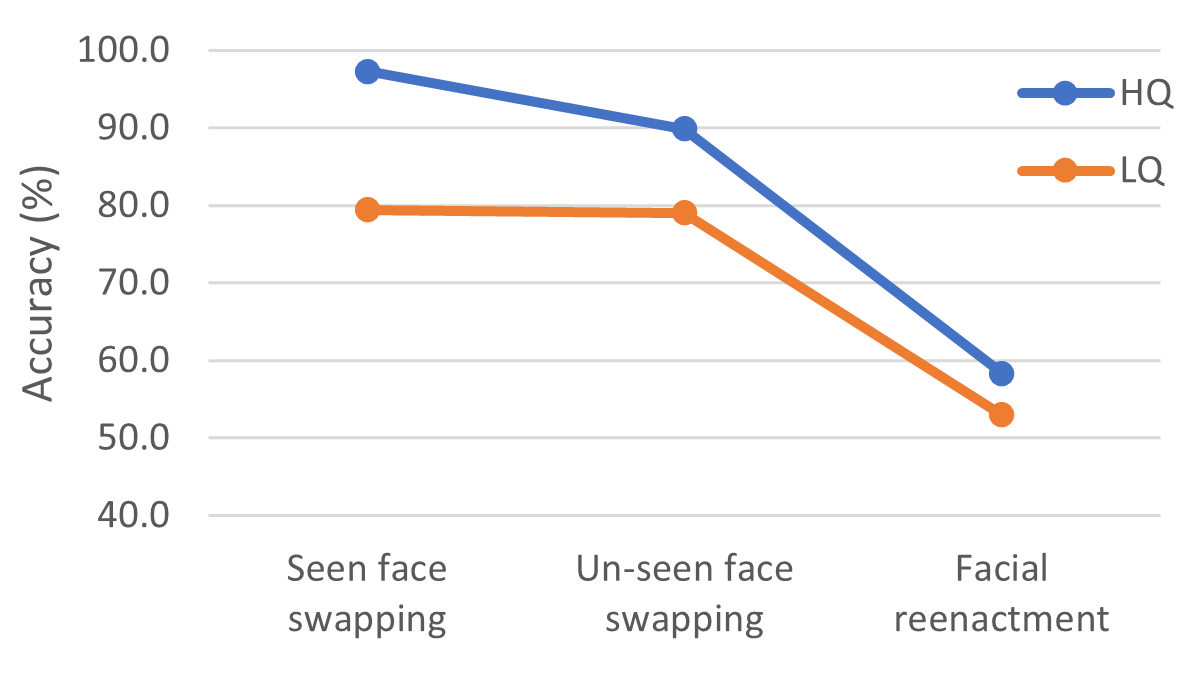}
    \caption{Accuracy results (binary classification task) of the winner of the Deepfake Detection Challenge \cite{Selimsef2020} trained on DFDC dataset \cite{Dolhansky2020dfdc} and tested on different datasets: the preview DFDC \cite{Dolhansky2019preview} (seen face swapping) and FaceForensics++ \cite{Roessler2019ff++} both on face swapping and facial-reenactment. Results are presented on high quality (HQ) and low quality (LQ) videos.}
    \label{fig:problem_explanation}
\end{figure}

To better highlight this problem, we carried out an experiment considering the winning solution of the recent DeepFake Detection Challenge organized by Facebook on Kaggle platform. 
The performers had the possibility to train their models using a huge dataset of videos (around 100k fake videos and 20k pristine ones with hundreds of different identities). 
In Fig.~\ref{fig:problem_explanation}, we show the results of our experiment.
The model was first tested on a dataset of real and deepfake videos including similar face-swapping manipulations, then we considered unseen face-swapping manipulations and finally videos manipulated using facial reenactment.
One can clearly observe the significant drop in performance in this last situation. 
Furthermore, the test on low quality compressed videos shows an additional loss and the final value for the accuracy is no more than a random guess.

It is also worth noting that current approaches are often used as black-box models and it is very difficult to predict the result because in a realistic scenario it is impossible to have a clue about the type of manipulation that occurred. 
The lack of reliability of current supervised deep learning methods pushed us to take a completely different perspective, avoiding to answer to a binary question (real or fake?) and instead focusing on wondering if the face under test preserves all the biometric traits of the involved subject.

Following this direction, our proposed method turns out to be able to generalize to different manipulation methods and also shows robustness w.r.t. low-quality data.
It can reveal the identity of a subject by highlighting inconsistencies of facial features such as temporal consistent motion.
The underlying CNN architecture comprises three main components:
a facial feature extractor, a temporal network to detect biometric anomalies (temporal ID network) and a generative adversarial network that tries to predict person-specific motion based on the expressions of a different subject.
The networks are trained only on real videos containing many different subjects \cite{Chung2018voxceleb2}.
During test time, in addition to the test video, we assume to have a set of pristine videos of the target person.
Based on these pristine examples, we compute a distance metric to the test video using the embedding of the temporal ID network (Fig.~\ref{fig:proposal_scheme}).
Overall, our main contributions are the following:
\begin{itemize}
\item We propose an example-based forgery detection approach that detects videos of facial manipulations based on the identity of the subject, especially the person-specific face motion.
\item An extensive evaluation that demonstrates the generalization to different types of manipulations even on low-quality videos, with a significant average improvement of more than $15$\% w.r.t. state of the art.
\end{itemize}

\section{Related Work}

Digital media forensics, especially, in the context of DeepFakes, is a very active research field.
The majority of the approaches rely on the availability of large-scale datasets of both pristine and fake videos for supervised learning.
A few approaches detect manipulations as anomalies w.r.t. features learned only on pristine videos.
Some of these approaches verify if the behavior of a person in a video is consistent with a given set of example videos of this person.
Our approach ID-Reveal is such an example-based forgery detection approach.
In the following, we discuss the most related detection approaches.

\paragraph{Learned features}

Afchar et al.~\cite{Afchar2018} presented one of the first approaches for DeepFake video detection based on supervised learning.
It focuses on mesoscopic features to analyze the video frames by using a network with a low number of layers.
R\"ossler et al.~\cite{Roessler2019ff++} investigated the performance of several CNNs architectures for DeepFake video detection and showed that very deep networks are more effective for this task, especially, on low-quality videos.
To train the networks, the authors also published a large-scale dataset.
The best performing architecture XceptionNet~\cite{Chollet2017} was applied frame-by-frame and has been further improved by follow-up works.
In \cite{Dang2020detecting} an attention mechanism is included, that can also be used to localize the manipulated regions, while in Kumar et al.~\cite{Kumar2020detecting} a triplet loss has been applied to improve performance on highly compressed videos.

Orthogonally, by exploiting artifacts that arise along the temporal direction it is possible to further boost performance.
To this end, Guera et al.~\cite{Guera2018} propose using a convolutional Long Short Term Memory (LSTM) network. Masi et al.~\cite{Masi2020two} propose to extract features by means of a two-branch network that are then fed into the LSTM: one branch takes the original
information, while the other one works on the residual image.
Differently in \cite{Zi2020WildDeepfake} a 3D CNN structure is proposed together with an attention mechanism at different abstraction levels of the feature maps.

Most of these methods achieve very good performance when the training set comprises the same type of facial manipulations, but performance dramatically impairs on unseen tampering methods.
Indeed, generalization represents the Achilles’ heel in media forensics.
Augmentation can be of benefit to generalize to different manipulations as shown in~\cite{Wang2020CNN}. In particular, augmentation has been extensively used by the best performing approaches during the DeepFake detection challenge~\cite{Dolhansky2020dfdc}.
Beyond the classic augmentation operations, some of them were particularly useful, e.g., by including cut-off based strategies on some specific parts of the face.
In addition to augmentation, ensembling different CNNs have been also used to improve performance during this challenge \cite{Bonettini2020video, Du2020towards}.
Another possible way to face generalization is to learn only on pristine videos and interpret a manipulation as an anomaly. This can improve the detection results on various types of face manipulations, even if the network never saw such forgeries in training.
In \cite{Cozzolino2019extracting} the authors extract the camera fingerprint information gathered from multiple frames and use those for detection.
Other approaches focus on specific operations used in current DeepFake techniques. For example, in \cite{Li2020face} the aim is to detect the blending operation that characterizes the face boundaries for most current synthetic face generation approaches.

A different perspective to improve generalization is presented in  \cite{Cozzolino2018, Aneja2020}, where few-shot learning strategies are applied. 
Thus, these methods rely on the knowledge of a few labeled examples of a new approach and guide the training process such that new embeddings can be properly separated from previous seen manipulation methods and pristine samples in a short retraining process.

\paragraph{Features based on physiological signals}

Other approaches look at specific artifacts of the generated videos that are related to physiological signals.
In \cite{Li2018ictu} it is proposed a method that detects eye blinking, which is characterized by a specific frequency and duration in videos of real people. 
Similarly, one can also use inconsistencies on head pose \cite{Yang2019head} or
face warping artifacts \cite{Li2019warping} as identifiers for tampered content.
Recent works are also using heart beat \cite{Fernandes2019, Qi2020DeepRhythm}
and other biological signals \cite{Ciftci2020FakeCatcher}
to find inconsistencies both in spatial and along the temporal direction.

\paragraph{Identity-based features}

The idea of identity-based approaches is to characterize each individual by extracting some specific biometric traits that can be hardly reproduced by a generator \cite{Agarwal2019, Agarwal2020detecting, Agarwal2020}.
The work by Agarwal et al. \cite{Agarwal2019} is the first approach that exploits the distinct patterns of facial and head movements of an individual to detect fake videos.
In \cite{Agarwal2020detecting} inconsistencies between the mouth shape dynamics and a spoken phoneme are exploited.
Another related work is proposed in \cite{Agarwal2020} to detect face-swap manipulations.
The technique uses both static biometric based on facial identity and temporal ones based on facial expressions and head movements. The method includes standard techniques from face recognition and a learned behavioral embedding using a CNN powered by a metric-learning objective function.
In contrast, our proposed method extracts facial features based on a 3D morphable model and focuses on temporal behavior through an adversarial learning strategy.
This helps to improve the detection of facial reenactment manipulations while still consistently be able to spot face swapping ones.
\begin{figure*}[t!]
    \centering
    \includegraphics[width=0.9\linewidth]{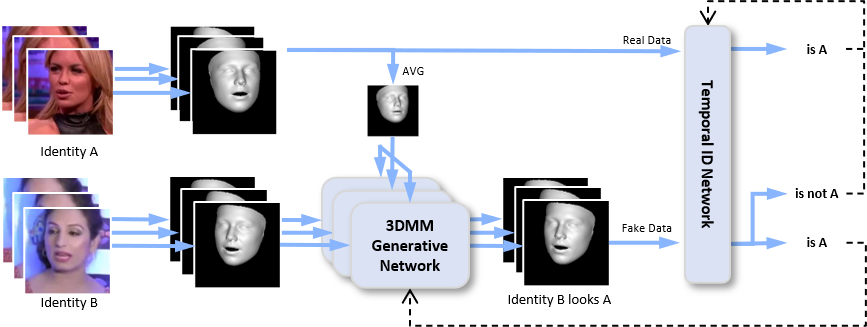}
    \caption{ID-Reveal is based on two neural networks, the \TemporalIDNetwork~ as well as the \GenNet, which interact with each other in an adversarial fashion.
    Using a three-dimensional morphable model (3DMM), we process videos of different identities and train the \TemporalIDNetwork~ to embed the extracted features such that they can be separated in the resulting embedding space based on their containing identity. In order to incentivize this network to focus on temporal aspects rather than visual cues, we jointly train the \GenNet~ to transform extracted features to fool its discriminative counterpart.
	}
    \label{fig:training}
\end{figure*}

\section{Proposed Method}
\label{sec:method}

ID-Reveal is an approach for DeepFake detection that uses prior biometric characteristics of a depicted identity, to detect facial manipulations in video content of the person.
Any manipulated video content based on facial replacement results in a disconnect between visual identity as well as biometrical characteristics.
While facial reenactment preserves the visual identity, biometrical characteristics such as the motion are still wrong.
Using pristine video material of a target identity we can extract these biometrical features and compare them to the characteristics computed on a test video that is potentially manipulated.
In order to be able to generalize to a variety of manipulation methods, we avoid training on a specific manipulation method, instead, we solely train on non-tampered videos.
Additionally, this allows us to leverage a much larger training corpus in comparison to the facial manipulation datasets~\cite{Roessler2019ff++,Dolhansky2020dfdc}.

Our proposed method consists of three major components (see Fig.~\ref{fig:training}).
Given a video as input, we extract a compact representation of each frame using a 3D Morphable model (3DMM)~\cite{Blanz1999}.
These extracted features are input to the \TemporalIDNetwork~ which computes an embedded vector.
During test time, a metric in the embedding space is used to compare the test video to the previously recorded biometrics of a specific person.
However, in order to ensure that the \TemporalIDNetwork~ is also based on behavioral instead of only visual information, we utilize a second network, called \GenNet, which is trained jointly in an adversarial fashion (using the \TemporalIDNetwork~as discriminator).
In the following, we will detail the specific components and training procedure.

\paragraph{Feature Extraction}
Our employed networks are based on per-frame extracted facial features.
Specifically, we utilize a low dimensional representation of a face based on a 3D morphable model~\cite{Blanz1999}.
The morphable model represents a 3D face by a linear combination of principle components for shape, expression, and appearance.
These components are computed via a principle component analysis of aligned 3D scans of human faces.
A new face can be represented by this morphable model, by providing the corresponding coefficients for shape, expression, and appearance.
To retrieve these parameters from video frames, one can use optimization-based analysis-by-synthesis approaches~\cite{thies2016face} or learned regression.
In our method, we rely on the regression framework of Guo et al.~\cite{Guo2020} which predicts a vector of $62$ coefficients for each frame.
Note that the $62$ parameters, contain $40$ coefficients for the shape, $10$ for the expression, and additional $12$ parameters for the rigid pose of the face (represented as a $3\times4$ matrix).
In the following, we denote the extracted 3DMM features of video $i$ of the individual $c$ at frame $t$ by $x_{c,i}(t) \in \mathbb{R}^{62}$.

\paragraph{Temporal ID Network}
The \TemporalIDNetwork~$\mathcal{N}_T$  processes the temporal sequence of 3DMM features through convolution layers that work along the temporal direction in order to extract the embedded vector $y_{c,i}(t) = \mathcal{N}_T \left[  x_{c,i}(t) \right]$.
To evaluate the distance between embedded vectors, we adopt the squared Euclidean distance, computing the following similarity:
\begin{equation}
    S_{c,i,k,j}(t) = -\frac{1}{\tau} \min_{t'} \left\| y_{c,i}(t) - y_{k,j}(t') \right\|^2
    \label{equ:sim}
\end{equation}
As a metric learning loss, similar to the Distance-Based Logistic Loss~ \cite{vo2016localizing}, we adopt a log-loss on a suitably defined probability~\cite{Cozzolino2020}.
Specifically, for each embedded vector $y_{c,i}(t)$, we build the probability through softmax processing as:
\begin{equation}
    p_{c,i}(t) = \frac{ \sum_{j\ne i} e^{S_{c,i,c,j}(t)} }{ \sum_{j\ne i} e^{S_{c,i,c,j}(t)} + \sum_{k\ne c} \sum_{j} e^{S_{c,i,k,j}(t)} },
    \label{equ:prob}
\end{equation}
Thus, we are considering all the similarities with respect to the pivot vector $y_{c,i}(t)$ in our probability definition $p_{c,i}(t)$.
Note that to obtain a high probability value it is only necessary that at least one similarity with the same individual is much larger than similarities with other individuals.
Indeed, the loss proposed here is a less restrictive loss compared to the current literature, where the aim is to achieve a high similarity for all the coherent pairs \cite{Hadsell2006, Hoffer2015, wang2019multi}.
The adopted metric learning loss is then obtained from the probabilities through the log-loss function:
\begin{equation}
    \mathcal{L}_{rec} =  \sum_{c,i,t} -log\left( p_{c,i}(t) \right).
    \label{equ:loss_rec}
\end{equation}
In order to tune hyper-parameters during training, we also measure the accuracy of correctly identifying a subject.
It is computed by counting the number of times where at least one similarity with the same individual is larger than all the similarities with other individuals.
The \TemporalIDNetwork~is first trained alone using the previously described loss, and afterward it is fine-tuned together with the \GenNet, which we describe in the following paragraph.

\paragraph{3DMM Generative Network}
The \GenNet~$\mathcal{N}_G$ is trained to generate 3DMM features similar to the features that we may extract from a manipulated video.
Specifically, the generative network has the goal to output features that are coherent to the identity of an individual, but with the expressions of another subject.
The generative network $\mathcal{N}_G$ works frame-by-frame and generates a 3DMM feature vector by combining two input feature vectors.
Let $x_c$ and $x_k$ are the 3DMM feature vectors respectively of the individuals $c$ and $k$, then, $\mathcal{N}_G \left[ x_k, x_c \right]$ is the generated feature vector with appearance of the individual $c$ and expressions of individual $k$.
During training, we use batches that contain $N\times M$ videos of $N$ different individuals each with $M$ videos.
In our experiments, we chose $M=N=8$.
To train the generative network $\mathcal{N}_G$, we apply it to pairs of videos of these $N$ identities.
Specifically, for each identity $c$, we compute an averaged 3DMM feature vector $\overline{x}_c$.
Based on this averaged input feature $\overline{x}_c$ and a frame feature $x_{i}(t)$ of a video of person $i$ (which serves as expression conditioning), we generate synthetic 3DMM features using the generator $\mathcal{N}_G$:
\begin{equation}
    x^*_{c,i}(t) = \mathcal{N}_G \left[ x_{i}(t), \overline{x}_c \right] .
\end{equation}

The \GenNet~is trained based on the following loss:
\begin{equation}
    \mathcal{L}_{\mathcal{N}_G} =  \mathcal{L}_{adv} + \lambda_{cycle} \, \mathcal{L}_{cycle}
    \label{equ:loss_gen}
\end{equation}
Where $\mathcal{L}_{cycle}$ is a cycle consistency used in order to preserve the expression.
Specifically, the \GenNet~is applied twice, firstly to transform a 3DMM feature vector of the individual $i$ to identity $c$ and then to transform the generated 3DMM feature vector to identity $i$ again, we should obtain the original 3DMM feature vector.
The $\mathcal{L}_{cycle}$ is defined as:
\begin{equation}
    \mathcal{L}_{cycle} = \sum_{c,i,t} \left\| x_{i}(t) - \mathcal{N}_G \left[ x^*_{c,i}(t), \overline{x}_i \right] \right\|^2 .
    \label{equ:loss_cycle}
\end{equation}
The adversarial loss $\mathcal{L}_{adv}$ is based on the \TemporalIDNetwork,
i.e., it tries to fool the \TemporalIDNetwork~by generating features that are coherent for a specific identity.
Since the generator works frame-by-frame, it can deceive the \TemporalIDNetwork~by only altering the appearance of the individual and not the temporal patterns.
The adversarial loss $\mathcal{L}_{adv}$ is computed as:
\begin{equation}
    \mathcal{L}_{adv} =  \sum_{c,i,t} -log\left( p^*_{c,i}(t) \right) , 
    \label{equ:loss_na}
\end{equation}
where the probabilities $p^*_{c,i}(t)$ are computed using the equation \ref{equ:prob}, but considering the similarities evaluated between generated features and real ones:
\begin{equation}
    S^*_{c,i,k,j}(t) = -\frac{1}{\tau} \min_{t'} \left\| \mathcal{N}_T\left[x^*_{c,i}(t) \right] - y_{k,j}(t') \right\|^2 .
\end{equation}
Indeed, the generator aims to increase the similarity between the generated features for a given individual and the real features of that individual.
During training, the \TemporalIDNetwork~is trained to hinder the generator, through a loss obtained as:
\begin{equation}
    \mathcal{L}_{\mathcal{N}_T} = \mathcal{L}_{rec} + \lambda_{inv} \mathcal{L}_{inv} .
    \label{equ:loss_ins}
\end{equation}
where the loss $\mathcal{L}_{inv}$, contrary to $\mathcal{L}_{abv}$, is used to minimized the probabilities $p^*_{c,i}(t)$.
Therefore, it is defined as:
\begin{equation}
    \mathcal{L}_{inv} =  \sum_{c,i,t} -log\left( 1 - p^*_{c,i}(t) \right) .
\end{equation}
Overall, the final objective of the adversarial game is to increase the ability of the \TemporalIDNetwork~to distinguish real identities from fake ones.

\paragraph{Identification}
Given a test sequence depicting a single identity as well as a reference set of pristine sequences of the same person, we apply the following procedure:
we first embed both the test as well as our reference videos using the \TemporalIDNetwork~pipeline.
We then compute the minimum pairwise Euclidean distance of each reference video and our test sequence.
Finally, we compare this distance to a fixed threshold $\tau_{id}$ to decide whether the behavioral properties of our testing sequence coincide with its identity, thus, evaluating the authenticity of our test video.
The source code and the trained network of our proposal are publicly available\footnote{\url{https://github.com/grip-unina/id-reveal}}.

\section{Results}
\label{sec:results}

To analyze the performance of our proposed method, we conducted a series of experiments.
Specifically, we discuss our design choices w.r.t. our employed loss functions and the adversarial training strategy based on an ablation study applied on a set of different manipulation types and different video qualities.
In comparison to state-of-the-art DeepFake video detection methods, we show that our approach surpasses these in terms of generalizability and robustness.

\begin{figure}[t!]
    \centering
    \includegraphics[width=0.85\linewidth, trim=175 0 175 0]{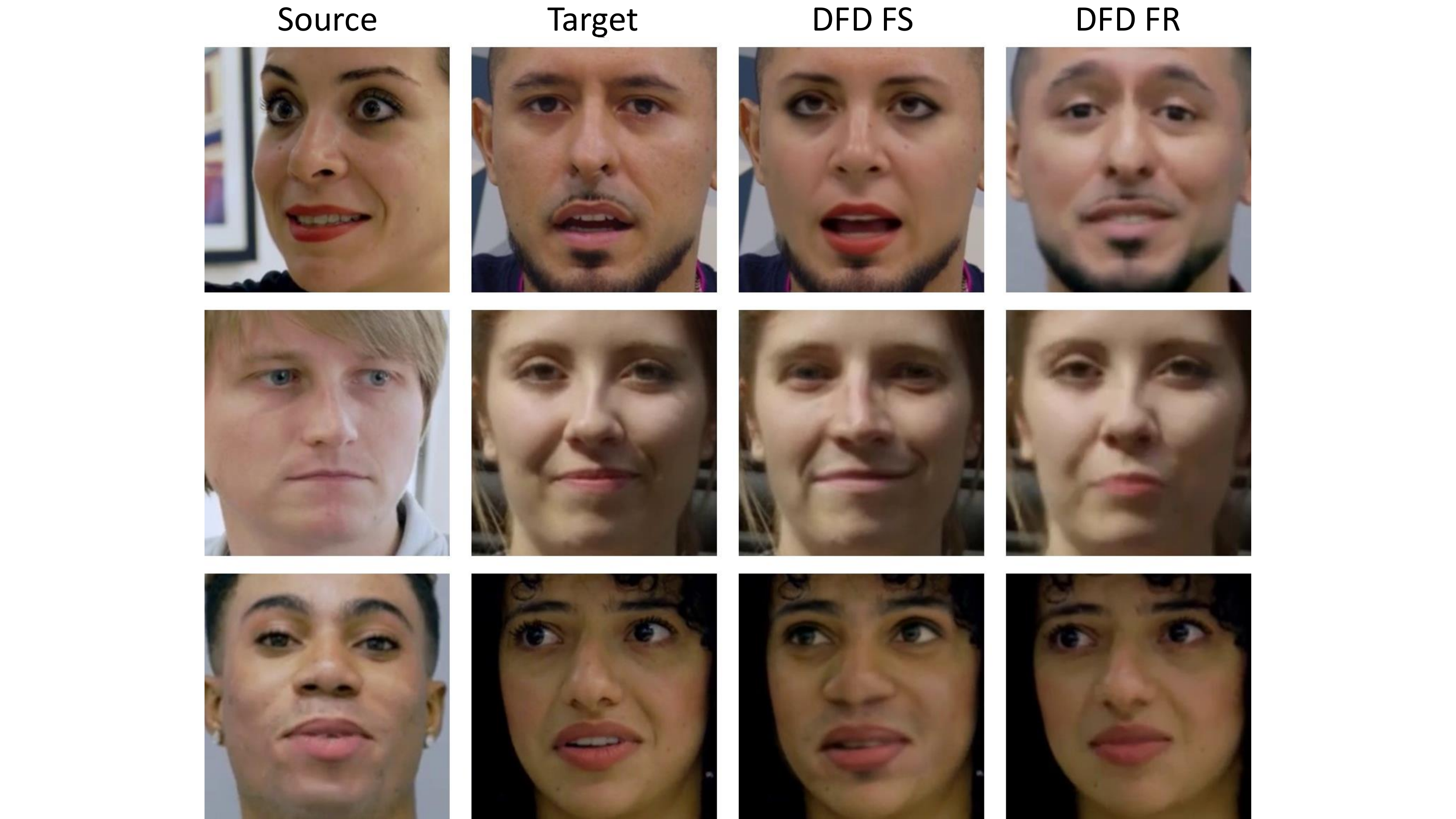}
    \caption{
        Aligned example images of DFD FS (Face-Swapping) as well as the newly created DFD FR (Facial Reenactment) datasets. From left to right: source videos, target sequences, DeepFakes and manipulations created using Neural Textures~\cite{thies2019}.
    }
    \label{fig:dfd_examples}
\end{figure}

\subsection{Experimental Setup}
Our approach is trained using the VoxCeleb2 development dataset \cite{Chung2018voxceleb2} consisting of multiple video clips of several identities.
Specifically, we use $5120$ subjects for the training-set and $512$ subjects for the validation-set.
During training, each batch contains $64$ sequences of $96$ frames.
The $64$ sequences are formed by $M=8$ sequences for each individual, with a total of $N=8$ different individuals extracted at random from the training-set.
Training is performed using the ADAM optimizer~\cite{adam}, with a learning rate of $10^{-4}$ and $10^{-5}$ for the Temporal ID Network and the 3DMM Generative Network, respectively.
The parameters $\lambda_{cycle}$, $\lambda_{inv}$ and $\tau$ for our loss formulation are set to $1.0$, $0.001$ and $0.08$ respectively.
We first train the Temporal ID Network for 300 epochs (with an epoch size of $2500$ iterations) and choose the  best performing model based on the validation accuracy.
Using this trained network, we enable our \GenNet\, and continue training for a fixed 100 epochs.
For details on our architectures, we refer to the supplemental document.
For all experiments, we use a fixed threshold of $\tau_{id}=\sqrt{1.1}$ to determine whether the behavioral properties of a test video coincide with those of our reference videos.
This threshold is set experimentally based on a one-time evaluation on $4$ real and $4$ fake videos from the original DFD~\cite{Dufour2019} using the averaged squared euclidean distance of real and manipulated videos.

\subsection{Ablation Study}
\label{sec:ablation}

In this section, we show the efficacy of the proposed loss and the adversarial training strategy.
For performance evaluation of our approach, we need to know the involved identity (the source identity for face-swapping manipulations and the target identity for facial reenactment ones).
Based on this knowledge, we can set up the pristine reference videos used to compute the final distance metric.
To this end, we chose a controlled dataset that includes several videos of the same identity, i.e., the recently created dataset of the Google AI lab, called DeepFake Dataset (DFD)~\cite{Dufour2019}. 
The videos contain $28$ paid actors in $16$ different contexts, furthermore,
for each subject there are pristine videos provided (varying from $9$ to $16$).
In total, there are $363$ real and $3068$ DeepFakes videos.
Since the dataset only contains face-swapping manipulations, we generated $320$ additional videos that include $160$ Face2Face~\cite{thies2016face} and $160$ Neural Textures~\cite{thies2019} videos.
Some examples are shown in Fig.~\ref{fig:dfd_examples}.

\begin{table}[t!]
\centering
{\footnotesize
    \begin{tabular}{rrccc}
\toprule
\multicolumn{2}{c}{\ru  Acc(\%) / AUC } & MSL & Triplet  & ours  \\ 
\midrule
\ru \multirow{ 4}{*}{\rotatebox[origin=c]{90}{w.o. adversarial\hspace{-0.1cm}}} & DFD~FR~HQ   & {\bf 73.8} / 0.83 & 73.6 / 0.85 & {\bf 73.8} / \bf{0.86} \\
\ru & ~~~~~~LQ    & 66.6 / 0.77 & 73.3 / 0.81 & {\bf 79.1} / \bf{0.87} \\ \cmidrule(lr){2-5}
\ru & DFD~FS~HQ   & {\bf 87.8} / 0.94 & 83.5 / 0.95 & 82.6 / \bf{0.96} \\ 
\ru & ~~~~~~LQ    & 74.0 / 0.92 & {\bf 77.0} / 0.93 & 73.0 / \bf{0.94} \\ \midrule 
\ru \multirow{ 4}{*}{\rotatebox[origin=c]{90}{w. adversarial}} & DFD~FR~HQ   & 68.9 / 0.81 & 71.6 / 0.85 & {\bf 75.6} / \bf{0.89} \\
\ru & ~~~~~~LQ    & 69.1 / 0.77 & 73.9 / 0.81 & {\bf 81.8} / \bf{0.90} \\ \cmidrule(lr){2-5}
\ru & DFD~FS~HQ   & {\bf 87.8} / 0.94 & 84.7 / 0.94 & 84.8 / \bf{0.96} \\
\ru & ~~~~~~LQ    & 78.0 / 0.92 & {\bf 78.9} / 0.91 & 78.1 / \bf{0.94} \\ \bottomrule 
    \end{tabular}
}
    \caption{Accuracy and AUC for variants of our approach. We compare three different losses: the multi-similarity loss (MSL), the triplet loss and our proposed loss (eq.\ref{equ:loss_rec}). In addition, we present the results obtained with and without the adversarial learning strategy on high quality (HQ) and low quality (LQ) videos manipulated using Facial Reenactment (FR) and Face Swapping (FS).}
    \label{tab:abl}
\end{table}

Performance is evaluated at video level using a leave-one-out strategy for the reference-dataset.
In detail, for each video under test, the reference dataset only contains pristine videos with a different context from the one under test.
The evaluation is done both on high quality (HQ) compressed videos (constant rate quantization parameter equal to 23) using H.264 and low quality (LQ) compressed videos (quantization parameter equal to 40).
This scenario helps us to consider a realistic situation, where videos are uploaded to the web, but also to simulate an attacker who further compresses the video to hide manipulation traces.
We compare the proposed loss to the triplet loss \cite{Hoffer2015} and the multi-similarity loss (MSL) \cite{wang2019multi}.
For these two losses, we adopt the cosine distance instead of the Euclidean one as proposed by the authors.
Moreover, hyper-parameters are chosen to maximize the accuracy to correctly identify a subject in the validation set.
Results for facial reenactment (FR) and face swapping (FS) in terms of Area Under Curve (AUC) and accuracy both for HQ and LQ videos are shown in Tab.~\ref{tab:abl}.
One can observe that our proposed loss gives a consistent improvement over the multi-similarity loss ($5.5$\% on average) and the triplet loss ($2.8$\% on average) in terms of AUC.
In addition, coupled with the adversarial training strategy performance, it gets better for the most challenging scenario of FR videos with a further improvement of around $3$\% for AUC and of $6$\% (on average) in terms of accuracy.

\subsection{Comparisons to State of the Art}

We compare our approach to several state-of-the-art DeepFake video detection methods.
All the techniques are compared using the accuracy at video level.
Hence, if a method works frame-by-frame, we average the probabilities obtained from 32 frames uniformly extracted from the video, as it is also done in \cite{Bonettini2020video, Afchar2018}.

\paragraph{State of the art approaches} The methods used for our comparison are
frame-based methods: MesoNet~\cite{Afchar2018}, Xception~\cite{Chollet2017}, FFD (Facial Forgery Detection)~\cite{Dang2020detecting}, Efficient-B7~\cite{Tan2019efficientnet};
ensemble methods: ISPL (Image and Sound Processing Lab)~\cite{Bonettini2020video}, Seferbekov~\cite{Selimsef2020}; 
temporal-based methods: Eff.B1 + LSTM,  ResNet + LSTM \cite{Guera2018} and
an Identity-based method: A\&B (Appearance and Behavior) \cite{Agarwal2020}. A detailed description of these approaches can be found in the supplemental document.
In order to ensure a fair comparison, all supervised approaches (frame-based, ensemble and temporal-based methods) are trained on the same dataset of real and fake videos,
while the identity-based ones (A\&B and our proposal) are trained instead on VoxCeleb2~\cite{Chung2018voxceleb2}.

\paragraph{Generalization and robustness analysis} To analyze the ability to generalize to different manipulation methods, training and test come from different datasets. Note that we will focus especially on generalizing from face swapping to facial reenactment. 

\setlength{\tabcolsep}{4.5pt}
\begin{table}[t!]
\centering
{\footnotesize
    \begin{tabular}{lcccc}
    \toprule 
\ru  & \multicolumn{2}{c}{High Quality (HQ)}  & \multicolumn{2}{c}{Low Quality (LQ)}  \\ \cmidrule(lr){2-3} \cmidrule(lr){4-5}
\ru Acc(\%) / AUC & DFD~~FR & DFD~~FS & DFD~~FR & DFD~~FS \\
    \midrule
\ru MesoNet       & 57.0 / 0.65 & 54.0 / 0.57 & 58.1 / 0.61 & 52.7 / 0.53 \\
\ru Xception      & 51.9 / 0.74 & 78.5 / 0.93 & 49.8 / 0.48 & 58.5 / 0.63 \\
\ru Efficient-B7  & 53.1 / 0.75 & 88.2 / 0.97 & 50.2 / 0.48 & 58.5 / 0.64 \\
\ru FFD           & 53.6 / 0.57 & 75.3 / 0.83 & 51.3 / 0.53 & 63.9 / 0.69 \\
\ru ISPL          & 61.4 / 0.71 & 85.2 / 0.93 & 53.9 / 0.55 & 64.9 / 0.72 \\
\ru Seferbekov    & 55.8 / 0.77 & \bf{91.8} / \bf{0.98} & 49.4 / 0.47 & 61.9 / 0.67 \\
\ru ResNet + LSTM & 52.2 / 0.56 & 60.0 / 0.65 & 56.1 / 0.62 & 58.7 / 0.64 \\
\ru Eff.B1 + LSTM & 53.6 / 0.72 & 86.6 / 0.95 & 50.9 / 0.57 & 61.6 / 0.76 \\ \cmidrule(lr){1-5}
\ru A\&B          & 74.1 / 0.78 & 75.6 / 0.77 & 59.5 / 0.60 & 63.2 / 0.61 \\
\ru ID-Reveal     & \bf{75.6} / \bf{0.87} & 84.8 / 0.96 & \bf{81.8} / \bf{0.90} & \bf{78.1} / \bf{0.94} \\
    \bottomrule
    \end{tabular}
}
\vspace{-3pt}
\caption{Video-level detection accuracy and AUC of our approach compared to state-of-the-art methods. Results are obtained on the DFD dataset on HQ videos and LQ ones, split in facial reenactment (FR) and face swapping (FS) manipulations. Training for supervised methods is carried out on DFDC, while for identity-based methods on VoxCeleb2.}
    \label{tab:resultsDFD}
\end{table}

In a first experiment we test all the methods on the DFD Google dataset that contains both face swapping and facial reenactment manipulations, as described in Section \ref{sec:ablation}.
In this case all supervised approaches are trained on DFDC~\cite{Dolhansky2020dfdc} with around 100k fake and 20k real videos. This is the largest DeepFake dataset publicly available and includes five different types of manipulations\footnote{\url{https://www.kaggle.com/c/deepfake-detection-challenge}}.
Experiments on (HQ) videos, with a compression factor of 23, and on low-quality (LQ) videos, where the factor is 40 are presented in terms of accuracy and AUC in Tab.~\ref{tab:resultsDFD}. 
Most methods suffer from a huge performance drop when going from face-swapping to facial-reenactment,
with an accuracy that often borders 50\%, equivalent to coin tossing.
The likely reason is that the DFDC training set includes mostly face-swapping videos, and methods with insufficient generalization ability are unable to deal with different manipulations.
This does not hold for ID-Reveal and A\&B, which are trained only on real data and, hence, have an almost identical performance with both types of forgeries.
For facial reenactment videos, this represents a huge improvement with respect to all competitors.
In this situation it is possible to observe a sharp performance degradation of most methods in the presence of strong compression (LQ videos).
This is especially apparent with face-swapping,  where some methods are very reliable on HQ videos but become almost useless on LQ videos. On the contrary, ID-Reveal suffers only a very small loss of accuracy on LQ videos, and outperforms all competitors, including A\&B, by a large margin. 

In another experiment, we use FaceForensics++~\cite{Roessler2019ff++} (HQ) for training the supervised methods, while the identity based methods are always trained on the VoxCeleb2 dataset~\cite{Chung2018voxceleb2}.
For testing, we use the preview DFDC Facebook dataset~\cite{Dolhansky2019preview} and CelebDF~\cite{Li2020celeb}. 
The preview DFDC dataset~\cite{Dolhansky2019preview} is composed only of face-swapping manipulations of 68 individuals.
For each subject there are 3 to 39 pristine videos with 3 videos for each context.
We consider 44 individuals which have at least 12 videos (4 contexts); obtaining a total of 920 real videos and 2925 fake videos.
CelebDF~\cite{Li2020celeb} contains 890 real videos and 5639 face-swapping manipulated videos. The videos are related to 59 individuals except for 300 real videos that do not have any information about the individual, hence, they cannot be included in our analysis. 
Results in terms of accuracy and AUC at video-level are shown in Tab.~\ref{tab:resultsADD}.
One can observe that also in this scenario our method achieves very good results for all the datasets, with an average improvement with respect to the best supervised approach of about 16\% on LQ videos. Even the improvement with respect to the identity based approach A\&B~\cite{Agarwal2020} is significant, around 14\% on HQ videos and 13\% on LQ ones.
Again the performance of supervised approaches worsens in unseen conditions of low-quality videos, while our method preserves its good performance.

\setlength{\tabcolsep}{4.5pt}
\begin{table}[t!]
\centering
{\footnotesize
    \begin{tabular}{lcccc}
    \toprule 
\ru  & \multicolumn{2}{c}{High Quality (HQ)}  & \multicolumn{2}{c}{Low Quality (LQ)}  \\ \cmidrule(lr){2-3} \cmidrule(lr){4-5}
\ru Acc(\%) / AUC & DFDCp & CelebDF  & DFDCp & CelebDF  \\
    \midrule
\ru MesoNet       & 53.6 / 0.74 & 50.1 / 0.75 & 51.9 / 0.67 & 50.1 / 0.67 \\
\ru Xception      & 72.0 / 0.79 &{\bf 77.2} / {\bf 0.88} & 59.9 / 0.61 & 55.0 / 0.58 \\
\ru Efficient-B7  & 71.8 / 0.78 & 71.4 / 0.80 & 57.3 / 0.62 & 51.3 / 0.56 \\
\ru FFD           & 63.1 / 0.69 & 69.2 / 0.76 & 51.6 / 0.55 & 56.4 / 0.59 \\
\ru ISPL          & 69.6 / 0.78 & 71.2 / 0.83 & 52.0 / 0.71 & 50.8 / 0.61 \\
\ru Seferbekov    & 72.0 / 0.85 & 75.3 / 0.86 & 54.0 / 0.63 & 54.8 / 0.62 \\
\ru ResNet + LSTM & 61.2 / 0.67 & 58.2 / 0.72 & 56.3 / 0.59 & 57.0 / 0.60 \\
\ru Eff.B1 + LSTM & 67.2 / 0.75 & 75.3 / 0.84 & 51.0 / 0.54 & 55.3 / 0.58 \\
\cmidrule(lr){1-5}
\ru A\&B          & 65.2 / 0.60 & 54.0 / 0.56 & 61.7 / 0.59 & 52.6 / 0.55 \\
\ru ID-Reveal     & {\bf 80.4} / {\bf 0.91} & 71.6 / 0.84 & {\bf 73.9} / {\bf 0.86} & {\bf 64.4} / {\bf 0.80} \\
    \bottomrule
    \end{tabular}
}
\vspace{-3pt}
\caption{Video-level detection accuracy and AUC of our approach compared to state-of-the-art methods. Results are obtained on DFDCp and CelebDF on HQ videos and LQ ones. Training for supervised methods is carried out on FF++, while for identity-based methods on VoxCeleb2.}
    \label{tab:resultsADD}
\end{table}

\begin{figure}[t!]
    \centering
    \includegraphics[width=0.97\linewidth,trim= 0 20 0 0]{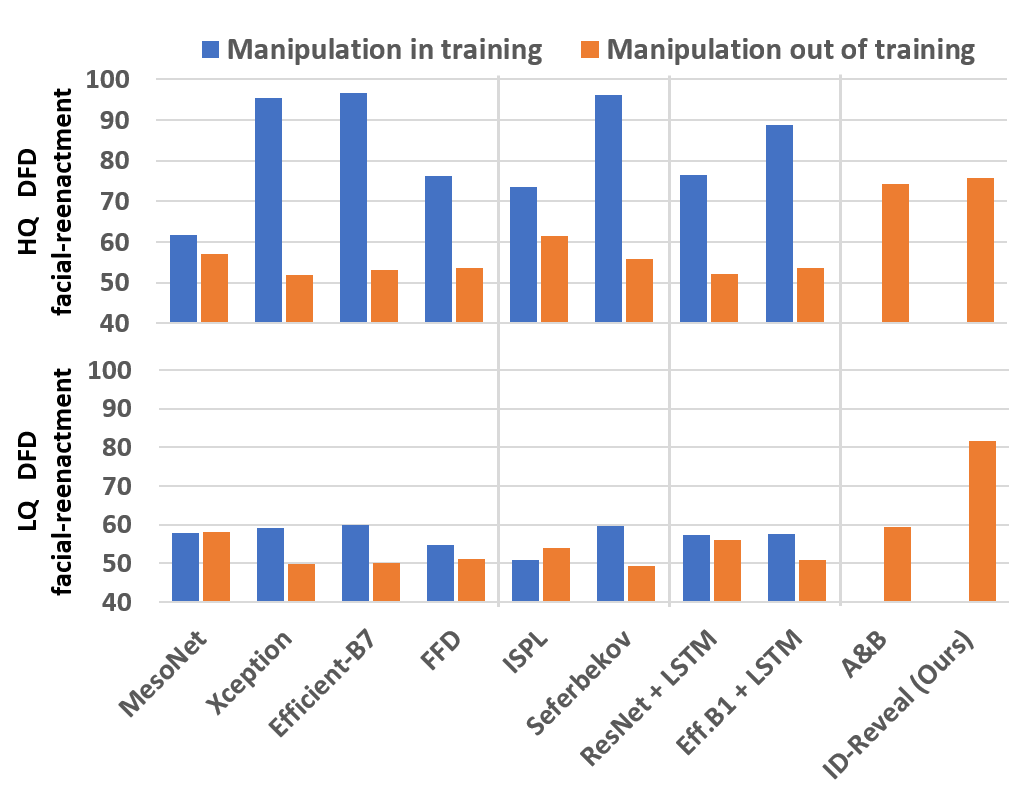}
    \caption{Binary detection accuracy of our approach compared to state-of-the-art methods. Results are obtained on the facial-reenactment DFD dataset both on HQ and LQ videos. We consider two different training scenarios for all the approaches that need forged videos in training:
    manipulation in training (blue bars), where the training set includes same type of manipulations present in test set (neural textures and face2face for facial reenactment), and  manipulation out of training (orange bars), where we adopt DFDC that includes only face swapping.}
    \label{fig:resultsDFDfr}
\end{figure}

To gain better insights on both generalization and robustness, we want to highlight the very different behavior of supervised methods when we change the fake videos in training. Specifically, for HQ videos if the manipulation (in this case neural textures and face2face) is included in training and test, then performance are very high for all the methods, but they suddenly decrease if we exclude those manipulations from the training, see Fig.~\ref{fig:resultsDFDfr}. The situation is even worse for LQ videos. Identity-based methods do not modify their performance since they do not depend at all on which manipulation is included in training. 

\section{Conclusion}
We have introduced ID-Reveal, an identity-aware detection approach leveraging a set of reference videos of a target person and trained in an adversarial fashion.
A key aspect of our method is the usage of a low-dimensional 3DMM representation to analyze the motion of a person.
While this compressed representation of faces contains less information than the original 2D images, the gained type of robustness is a very important feature that makes our method generalize across different forgery methods.
Specifically, the 3DMM representation is not affected by different environments or lighting situations, and is robust to disruptive forms of post-processing, e.g., compression.
We conducted a comprehensive analysis of our method and in comparison to state of the art, we are able to improve detection qualities by a significant margin, especially, on low-quality content.
At the same time, our method improves generalization capabilities by adopting a training strategy that solely focuses on non-manipulated content.

\section*{Acknowledgment}

We gratefully acknowledge the support of this research by a TUM-IAS Hans Fischer Senior Fellowship, a TUM-IAS Rudolf M\"o{\ss}bauer Fellowship and a Google Faculty Research Award. In addition, this material is based on research sponsored by the Defense Advanced Research Projects Agency (DARPA) and the Air Force Research Laboratory (AFRL) under agreement number FA8750-20-2-1004. 
The U.S. Government is authorized to reproduce and distribute reprints for Governmental purposes notwithstanding any copyright notation thereon. 
The views and conclusions contained herein are those of the authors and should not be interpreted as necessarily representing the official policies or endorsements, either expressed or implied, of DARPA and AFRL or the U.S. Government. This work is also supported by the PREMIER project, funded by the Italian Ministry of Education, University, and Research within the PRIN 2017 program.

\begin{figure*}[t]
    \centering
    \includegraphics[width=0.88\linewidth]{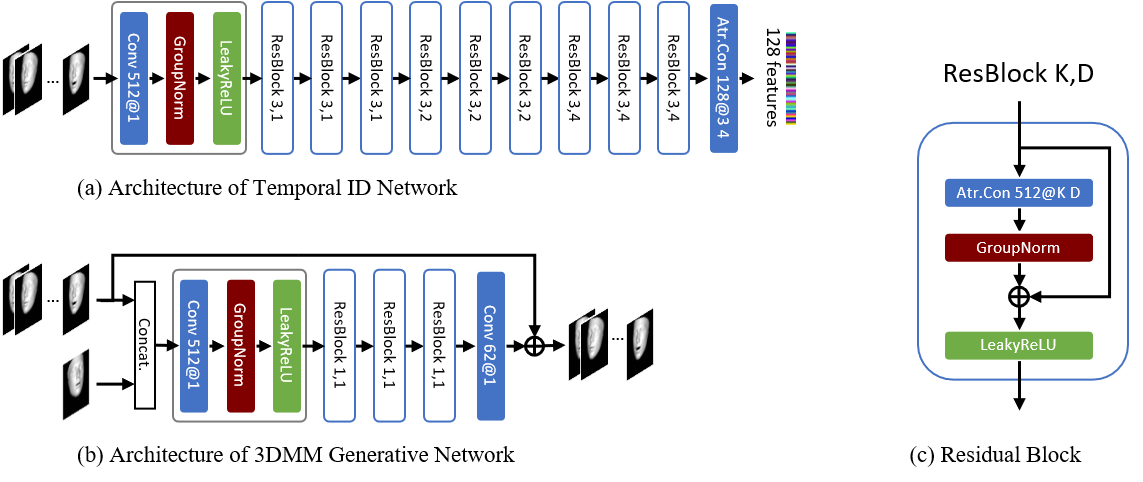}
    \caption{Architecture of our proposed Temporal ID Network and 3DMM Generative Network.}
    \label{fig:arc_nt}
\end{figure*}
\begin{appendix}
\section*{Appendix}
In this appendix, we report the details of our architectures used for the \TemporalIDNetwork~and the \GenNet~(Sec.~\ref{sec:architectures}).
Moreover, we briefly describe the state of the art DeepFake methods we compare to, 
(see Sec.~\ref{sec:sota}).
In Sec.~\ref{sec:addition} and Sec.~\ref{sec:contexts}, we present additional results to prove the generalization capability of our method.
In Sec.~\ref{sec:visualization}, we include scatter plots that show the separability of videos of different subjects in the embedding space.
Finally, we analyze a real case on the web (see Sec.~\ref{sec:realcase}).

\section{Architectures}

\label{sec:architectures}

\paragraph{\TemporalIDNetwork} 
We leverage a convolution neural network architecture that works along the temporal direction and is composed by eleven layers (see Fig.\ref{fig:arc_nt} (a)).
We use Group Normalization \cite{Wu2018Group} and LeakyReLU non-linearity for all layers except the last one. 
Moreover, we adopt à-trous convolutions (also called dilated convolution), instead of classic convolutions in order to increase the receptive fields without increasing the trainable parameters.
The first layer increases the number of channels from 62 to 512, while the successive ones are inspired by the ResNet architecture \cite{He2016Deep} and include a residual block as shown in Fig.\ref{fig:arc_nt} (c).
The parameters $K$ and $D$ of the residual blocks are the dimension of the filter and the dilatation factor of the à-trous convolution, respectively.
The last layer reduces the channels from 512 to 128.
The receptive field of the whole network is equal to 51 frames which is around 2 seconds.

\paragraph{\GenNet} 
As described in the main paper, the \GenNet~is fed by two 3DMM feature vectors.
The two feature vectors are concatenated which results in a single input vector of 124 channels.
The network is formed by five layers: a layer to increase the channels from 124 to 512, three residual blocks, and a last layer to decrease the channels from 512 to 62. 
The output is summed to the input 3DMM feature vector to obtain the generated 3DMM feature vector (see Fig.\ref{fig:arc_nt} (b)).
All the convolutions have a dimension of filter equal to one in order to work frame-by-frame.

\section{Comparison methods}
\label{sec:sota}
In the main paper, we compare our approach with several state of the art DeepFake detection methods, that are described in following:

\vspace{-0.25cm}
\paragraph{Frame-based methods}
\begin{enumerate}[label=(\roman*)]
    \vspace{-0.15cm}
    \item MesoNet~\cite{Afchar2018}: is one of the first CNN methods proposed for DeepFake detection which uses dilated convolutions with inception modules.
    
    \vspace{-0.25cm}
    \item Xception~\cite{Chollet2017}: is a relatively deep neural network that is achieving a very good performance compared to other CNNs for video DeepFake detection~\cite{Roessler2019ff++}.  

    \vspace{-0.25cm}
    \item FFD (Facial Forgery Detection)~\cite{Dang2020detecting}: is a variant of Xception, including an attention-based layer, in order to focus on high-frequency details.

    \vspace{-0.25cm}
    \item Efficient-B7~\cite{Tan2019efficientnet}: has been proposed by Tan et al. and is pre-trained on ImageNet using the strategy described in \cite{Xie2020}, where the network is trained with injected noise (such as dropout, stochastic depth, and data augmentation) on both labeled and unlabeled images.
    
\end{enumerate}
\vspace{-0.25cm}
\paragraph{Ensemble methods}

\begin{enumerate}[label=(\roman*)]
    \setcounter{enumi}{4}
    \vspace{-0.15cm}
    \item ISPL (Image and Sound Processing Lab)~\cite{Bonettini2020video}: employs an ensemble of four variants of Efficienet-B4. The networks are trained using different strategies, such as self-attention mechanism and triplet siamese strategy. Data augmentation is performed by applying several operations, like downscaling, noise addition and JPEG compression. 

    \vspace{-0.25cm}
    \item Seferbekov~\cite{Selimsef2020}: is the algorithm proposed by the winner of the Kaggle competition (Deepfake Detection Challenge) organized by Facebook \cite{Dolhansky2020dfdc}. It uses an ensemble of seven Efficientnet-B7 that work frame-by-frame. The networks are pre-trained using the strategy described in \cite{Xie2020}. The training leverages data augmentation, that, beyond some standard operations, includes a cut-out that drops specific parts of the face.
\end{enumerate}

\vspace{-0.25cm}
\paragraph{Temporal-based methods}

\begin{enumerate}[label=(\roman*)]
    \setcounter{enumi}{6}
    \vspace{-0.15cm}
    \item ResNet + LSTM: is a method based on Long Short Term Memory (LSTM) \cite{Guera2018}. In detail, a ResNet50 is used to extract frame-level features from 20 frames uniformly extracted from the video. These features are provided to a LSTM that classifies the whole video.

    \vspace{-0.25cm}
    \item Eff.B1 + LSTM: This is a variant of the approach described above, where the ResNet architecture is replaced by EfficientNet-B1.

\end{enumerate}

\setlength{\tabcolsep}{4.5pt}
\begin{table}[t!]
\centering
{\footnotesize
    \begin{tabular}{llcccc}
    \toprule 
\multicolumn{2}{c}{\ru} & \multicolumn{2}{c}{High Quality (HQ)}  & \multicolumn{2}{c}{Low Quality (LQ)}  \\ \cmidrule(lr){3-4} \cmidrule(lr){5-6}
\multicolumn{2}{c}{\ru Acc(\%) / AUC} & DFD~~FR & DFD~~FS & DFD~~FR & DFD~~FS \\
    \midrule
\ru MesoNet   & Mean & 57.0 / 0.65 & 54.0 / 0.57 & 58.1 / 0.61 & 52.7 / 0.53 \\
\ru           & Max  & 54.5 / 0.55 & 52.6 / 0.47 & 54.2 / 0.55 & 51.6 / 0.48 \\
\ru Xception  & Mean & 51.9 / 0.74 & 78.5 / 0.93 & 49.8 / 0.48 & 58.5 / 0.63 \\
\ru           & Max  & 58.9 / 0.71 & 80.4 / 0.92 & 46.1 / 0.44 & 51.6 / 0.59 \\
\ru Effic.-B7 & Mean & 53.1 / 0.75 & 88.2 / 0.97 & 50.2 / 0.48 & 58.5 / 0.64 \\
\ru           & Max  & 62.5 / 0.73 & 79.4 / 0.96 & 45.2 / 0.44 & 55.9 / 0.66 \\
\ru FFD       & Mean & 53.6 / 0.57 & 75.3 / 0.83 & 53.9 / 0.55 & 64.9 / 0.72 \\
\ru           & Max  & 52.5 / 0.56 & 60.2 / 0.78 & 50.9 / 0.50 & 51.7 / 0.64 \\
    \midrule
\ru AVG       & Mean & 53.9 / 0.68 & 74.0 / 0.83 & 53.0 / 0.53 & 58.7 / 0.63 \\
\ru           & Max  & 57.1 / 0.64 & 68.2 / 0.78 & 49.1 / 0.48 & 52.7 / 0.59 \\
    \bottomrule
    \end{tabular}
}
\vspace{-3pt}
\caption{Video-level detection accuracy and AUC of frame-based methods. We compare two strategies: averaging the score over 32 frames in a video and taking the maximum score. Results are obtained on the DFD dataset on HQ videos and LQ ones, split in facial reenactment (FR) and face swapping (FS) manipulations.}
\label{tab:resultsDFD_max}
\end{table}

\vspace{-0.25cm}
\paragraph{Identity-based methods}
\begin{enumerate}[label=(\roman*)]
    \setcounter{enumi}{8}
    \vspace{-0.15cm}
    \item A\&B (Appearance and Behavior) \cite{Agarwal2020}:
    is an identity-based approach that includes a face recognition network and a network that is based on head movements.
    The behavior recognition system encodes the information about the identity through a network that works on a sequence of attributes related to the movement \cite{Wiles18fab}. 
\end{enumerate}

\noindent
Note that all the techniques are compared at video level.
Hence, if a method works frame-by-frame, we average the probabilities obtained from 32 frames uniformly extracted from the video.
Furthermore, to validate this choice, we compare averaging with the maximum strategy. Results are reported in Tab.~\ref{tab:resultsDFD_max} using the same experimental setting of Tab. 2 of the main paper.
The results prove the advantage to use the averaging operation with respect to the maximum value:
the increase in terms of AUC is around 0.04, while the accuracy increases (on average) of about 3\%.

\setlength{\tabcolsep}{4.5pt}
\begin{table}[t!]
\centering
{\footnotesize
    \begin{tabular}{lcccc}
    \toprule 
\ru & \multicolumn{2}{c}{High Quality (HQ)}  & \multicolumn{2}{c}{Low Quality (LQ)}  \\ \cmidrule(lr){2-3} \cmidrule(lr){4-5}
\ru Acc(\%) / AUC & FF++~FR & FF++~FS & FF++~FR & FF++~FS \\
    \midrule
\ru MesoNet          & 55.4 / 0.58 & 57.1 / 0.61  & 55.4 / 0.57 & 57.3 / 0.62 \\
\ru Xception         & 55.6 / 0.58 & 79.0 / 0.89  & 51.9 / 0.57 & 69.2 / 0.79 \\
\ru Efficient-B7     & 54.9 / 0.59 & 85.4 / 0.93  & 50.6 / 0.54 & 65.6 / 0.80 \\
\ru FFD              & 54.4 / 0.56 & 69.2 / 0.75  & 53.5 / 0.56 & 63.3 / 0.70 \\
\ru ISPL             & 56.6 / 0.59 & 74.2 / 0.83  & 53.3 / 0.55 & 68.8 / 0.76 \\
\ru Seferbekov       & 58.3 / 0.62 & 89.9 / 0.97  & 53.0 / 0.55 & 79.4 / 0.87 \\
\ru ResNet + LSTM    & 55.0 / 0.58 & 59.0 / 0.63  & 56.2 / 0.58 & 61.9 / 0.66 \\
\ru Eff.B1 + LSTM    & 57.2 / 0.62 & 81.8 / 0.90  & 54.1 / 0.58 & 69.0 / 0.78 \\ \cmidrule(lr){1-5}
\ru A\&B             & 72.2 / 0.78 & 89.0 / 0.97  & 51.5 / 0.53 & 51.9 / 0.65 \\
\ru ID-Reveal (Ours) & 78.3 / 0.87 & 93.6 / 0.99  & 74.8 / 0.83 & 81.9 / 0.97 \\
    \bottomrule
    \end{tabular}
}
\vspace{-3pt}
\caption{Video-level detection accuracy and AUC of our approach compared to state-of-the-art methods. Results are obtained on the FF++ dataset on HQ videos and LQ ones, split in facial reenactment (FR) and face swapping (FS) manipulations. Training for supervised methods is carried out on DFDC, while for identity-based methods on VoxCeleb2.}
    \label{tab:resultsFFpp}
\end{table}

\section{Additional results}
\label{sec:addition}

To show the ability of our method to be agnostic to the type of manipulation, we test our proposal on additional datasets, that are not included in the main paper.
In Tab.~\ref{tab:resultsFFpp} we report the analysis on the dataset FaceForensics++ (FF++) \cite{Roessler2019ff++}. Results are split for facial reenactment (FR) and face swapping (FS) manipulations.
It is important to underline that this dataset does not provide information about multiple videos of the same subject, therefore, for identity-based approaches, the first 6 seconds of each pristine video are used as reference dataset, while the last 6 seconds are used to evaluate the performance (we only consider videos of at least 14 seconds duration, thus, obtaining 360 videos for each manipulation method).
For the FF++ dataset, our method obtains always better performance in both the cases of  high-quality videos and low-quality ones.

\begin{figure}[t!]
    \centering
    \includegraphics[width=0.98\linewidth, trim=100 0 110 0, clip]{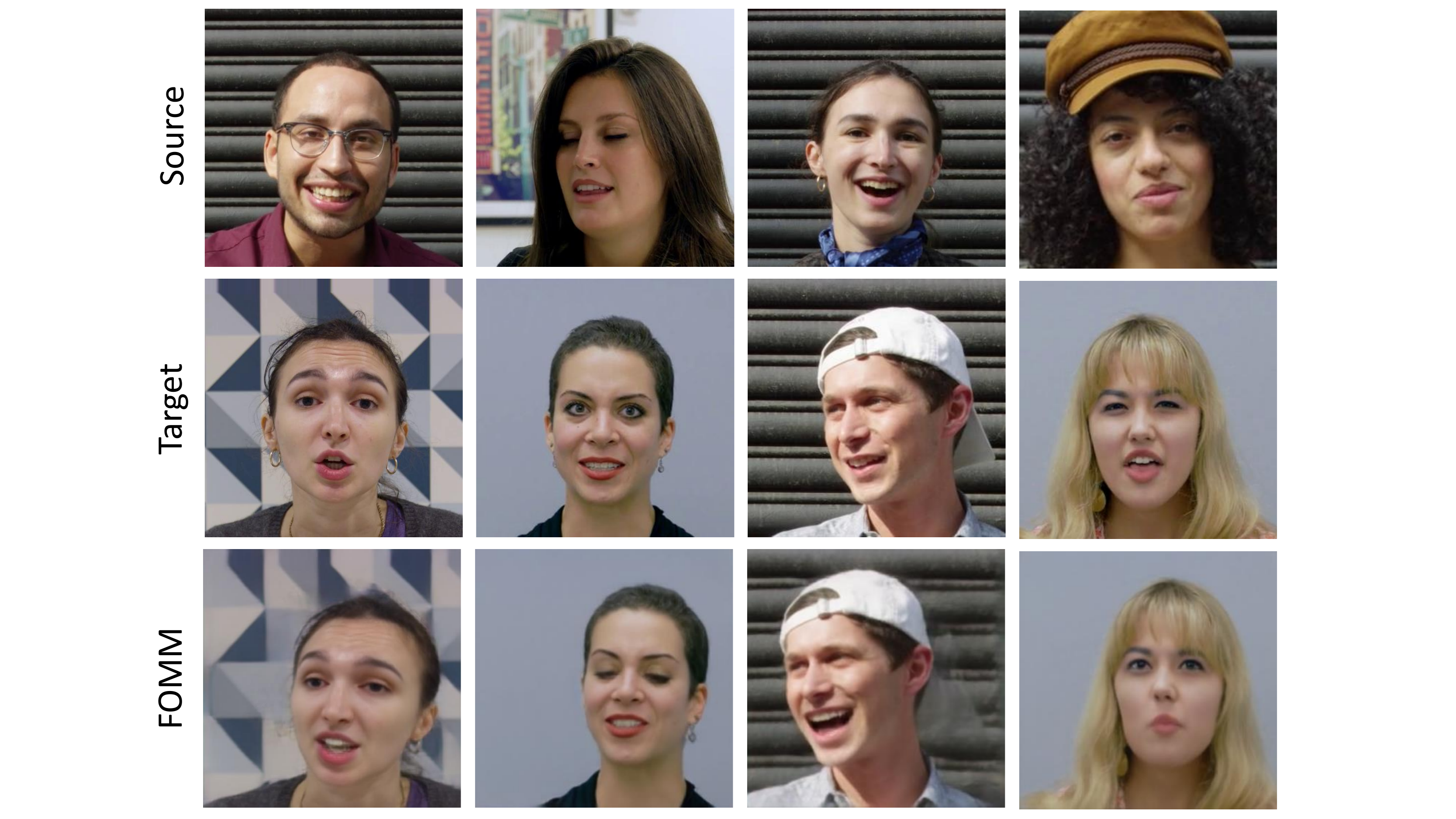}
    \caption{
        Aligned examples of created FOMM videos. From top to bottom: source videos, target sequences, and manipulations created using First-Order Motion Model \cite{Siarohin2019first}.
    }
    \label{fig:fomm_examples}
\end{figure}

As a further analysis, we test our method on a recent method of face reenactment, called  FOMM (First-Order Motion Model) \cite{Siarohin2019first}. Using the official code of FOMM, we created 160 fake videos using the pristine videos of DFD, some examples are in Fig.~\ref{fig:fomm_examples}. Our approach on these videos achieves an accuracy of 85.6\%, and an AUC of 0.94 which further underlines the generalization of our method with respect to a new type of manipulation.

\section{Robustness to different contexts}
\label{sec:contexts}

\begin{figure}[t!]
    \centering
    \includegraphics[width=1.0\linewidth, page=1, trim= 0 170 70 -10]{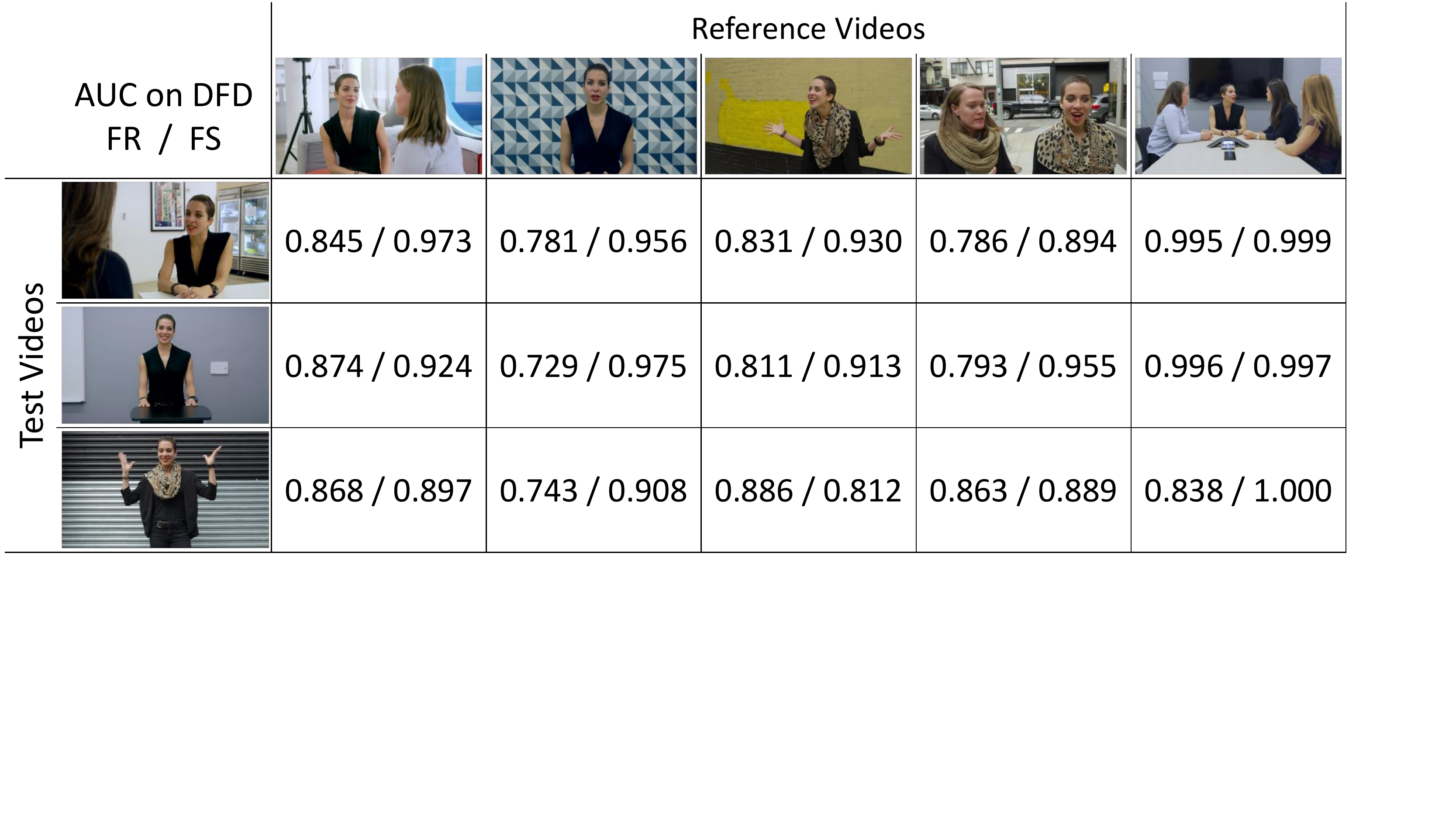}
    \caption{Average performance in terms of AUC evaluated on 28 actors of DFD FR and DFD FS datasets when test videos are in different contexts with respect to reference videos.
    Test videos: kitchen, podium-speech, outside laughing talking. Reference videos: angry talking, talking against wall, outside happy hugging, outside surprised, serious meeting.
 }
    \label{fig:context}
\end{figure}

We made additional experiments to understand that for our method it is not necessary that the reference videos are similar to the manipulated ones in terms of environment, lighting, or distance from the subject.
To this end, we show results in Fig.~\ref{fig:context} obtained for the DFD FR and DFD FS datasets,
where information about the video context (kitchen, podium, outside, talking, meeting, etc.) is available.
While the reference videos and the under-test videos differ, our method shows robust performance.
Results seem only affected by the variety of poses and expressions present in the reference videos (the last reference video in the table contains the most variety in motion, thus yielding better results). 

\section{Visualization of the embedded vectors}
\label{sec:visualization}

\begin{figure}[b!]
    \centering
    \includegraphics[width=0.49\linewidth]{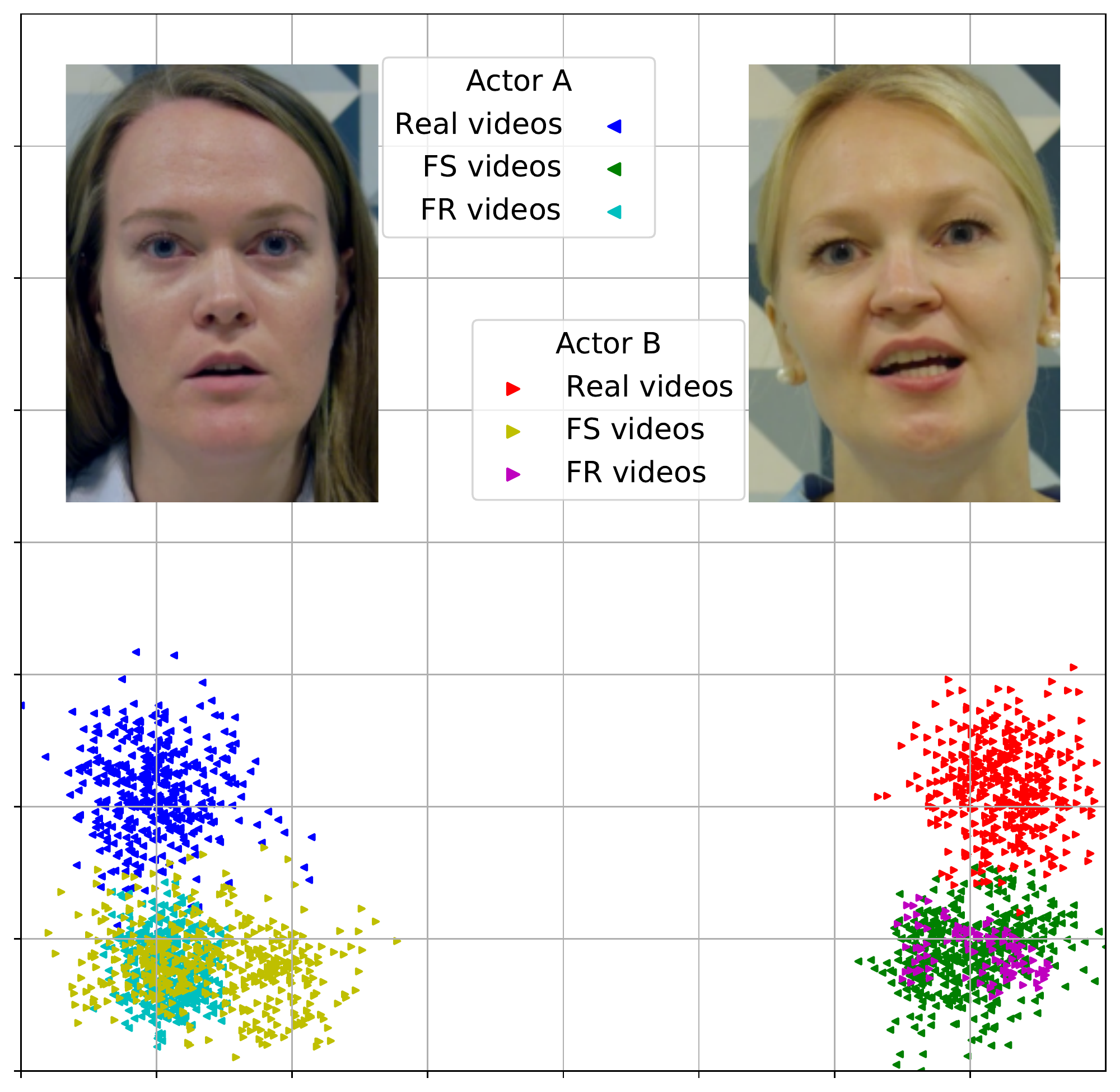}
    \includegraphics[width=0.49\linewidth]{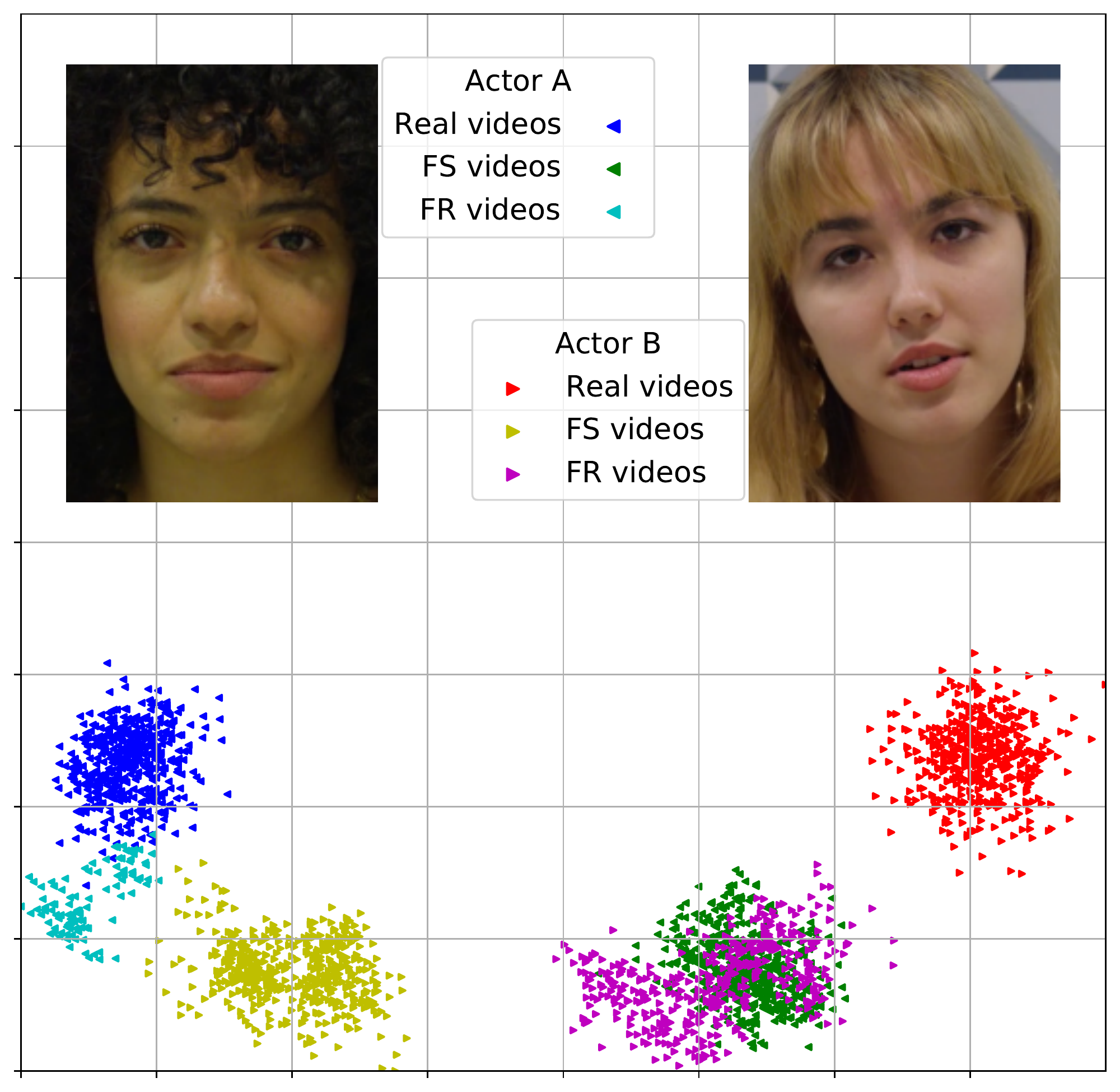}
    \includegraphics[width=0.49\linewidth]{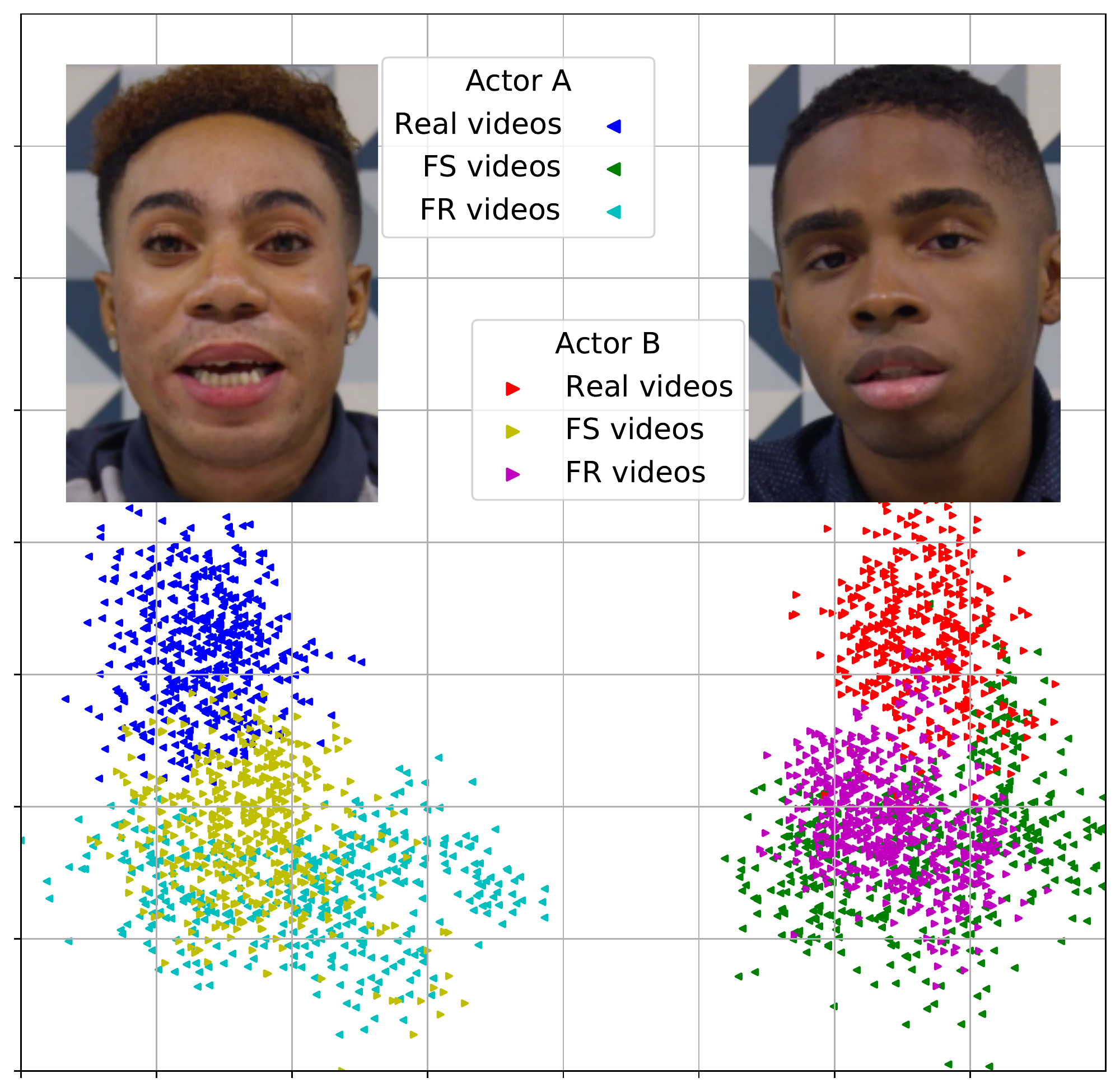}
    \includegraphics[width=0.49\linewidth]{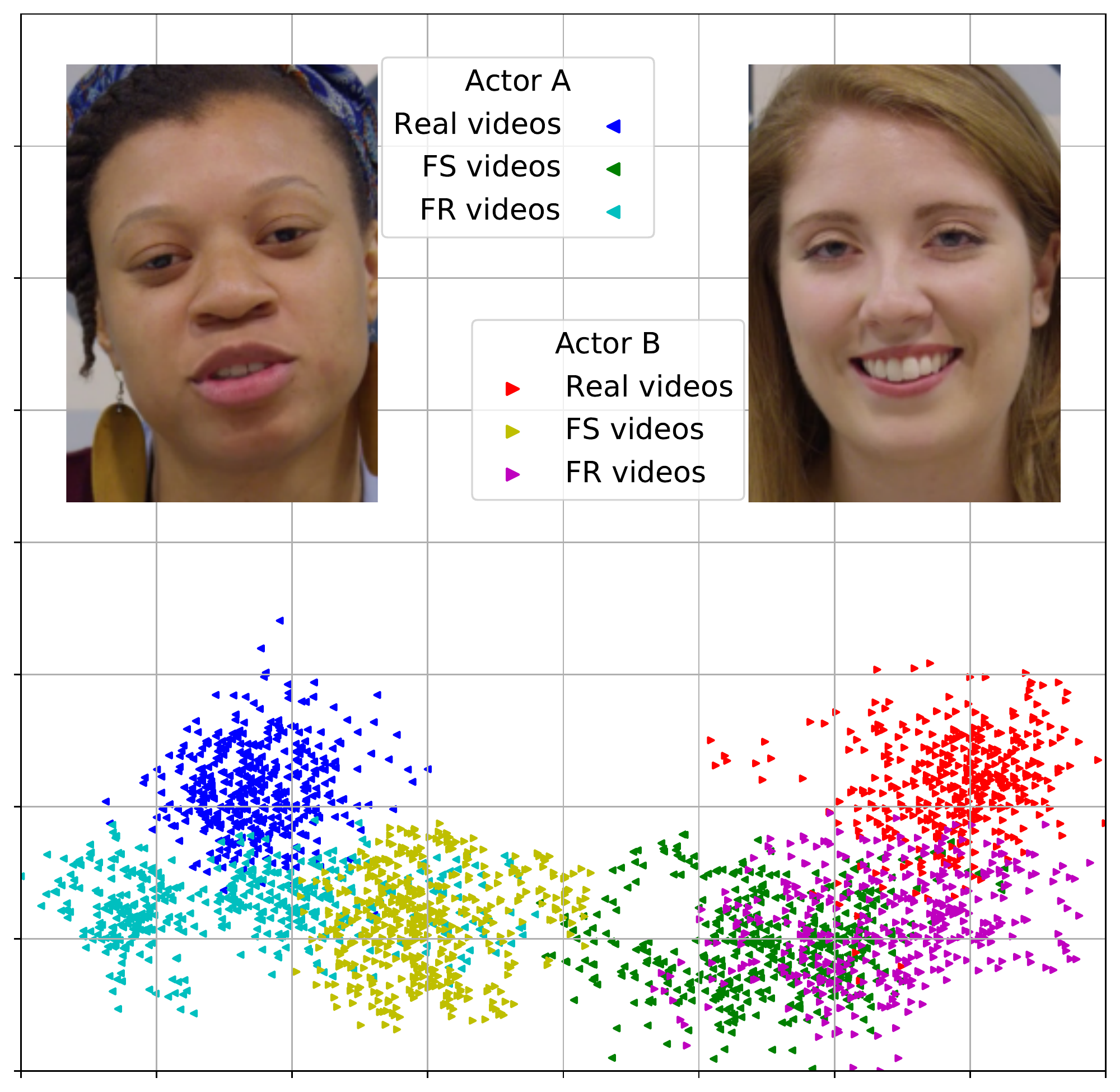}
    \caption{Scatter plots of embedded vectors extracted from 4 seconds long video snippets relative to couple of actors. We included both face swapping (FS) and facial reenactment (FR).}
    \label{fig:visualization}
\end{figure}

In this section, we include scatter plots that show the 2D orthogonal projection of the extracted temporal patterns. In particular, in Fig.~\ref{fig:visualization} we show
the scatter plots of embedded vectors extracted from 4 seconds long video snippets relative to two actors for the DFD dataset by using Linear Discriminant Analysis (LDA) and selecting the 2-D orthogonal projection that maximize the separations between the real videos of two actors and between real videos and fake ones.
We can observe that in the embedding space the real videos relative to different actors are perfectly separated. Moreover, also the manipulated videos relative to an actor are well separated from the real videos of the same actor. 

\begin{figure*}[t!]
    \centering
    \includegraphics[width=1.0\linewidth]{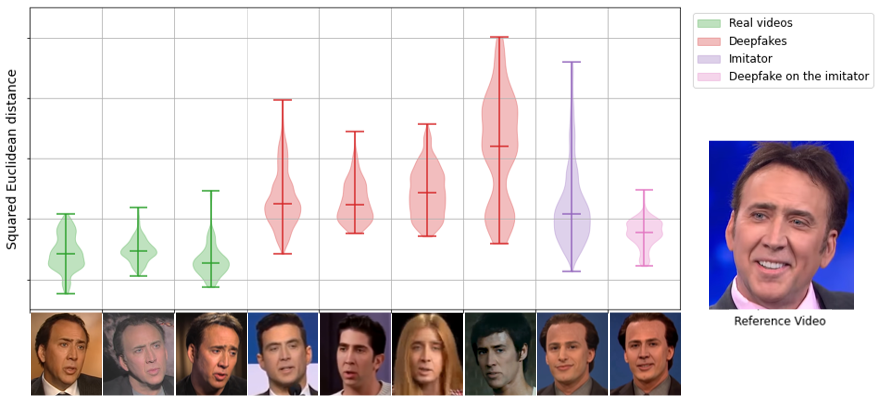}
    \caption{Distributions of squared Euclidean distances of 9 videos downloaded from YouTube with respect to a real reference video of Nicolas Cage. From left to right: 3 real videos, 4 DeepFakes, a video from an imitator and a DeepFake driven by the imitator.}
    \label{fig:youtube}
\end{figure*}

\section{A real case on the web}
\label{sec:realcase}

We applied ID-Reveal to videos of Nicolas Cage downloaded from YouTube.
We tested on three real videos, four DeepFakes videos, one imitator (a comic interpreting Nicolas Cage) and a DeepFake applied on the imitator.
We evaluate the distributions of distance metrics that are computed as the minimum pairwise squared Euclidean distance in the embedding space of 4 seconds long video snippets extracted from the pristine reference video and the video under test.
In Fig.~\ref{fig:youtube}, we report these distributions using a violin plot.

We can observe that the lowest distances are relative to real videos (green).
For the DeepFakes (red) all distances are higher and, thus, can be detected as fakes.
An interesting case is the video related to the imitator (purple), that presents a much lower distance since he is imitating Nicolas Cage.
A DeepFake driven by the imitator strongly reduces the distance (pink), but is still detected by our method.

\end{appendix}

\balance
{\small
\bibliographystyle{ieee_fullname}
\bibliography{egbib}
}

\end{document}